\documentclass[10pt,conference]{IEEEtran}

\usepackage{cite}

\ifCLASSINFOpdf
   \usepackage[pdftex]{graphicx}
   \graphicspath{{figs/}}
   \DeclareGraphicsExtensions{.pdf,.jpeg,.png}
\else
   \usepackage[dvips]{graphicx}
   \graphicspath{{../figs/}}
   \DeclareGraphicsExtensions{.eps}
\fi
\usepackage[cmex10]{amsmath}
\usepackage{amsmath,amsfonts}

\usepackage{booktabs}
\usepackage{bm}
\usepackage{color}

\usepackage{array}

\ifCLASSOPTIONcompsoc
  \usepackage[caption=false,font=normalsize,labelfont=sf,textfont=sf]{subfig}
\else
  \usepackage[caption=false,font=footnotesize]{subfig}
\fi
\usepackage{url}

\begin{document}
%
\title{From Actions to Events: A Transfer Learning Approach Using Improved Deep Belief Networks} 
\newif\iffinal
\finalfalse
\finaltrue
\newcommand{\cmtid}{XXX}


\iffinal

\author{\IEEEauthorblockN{Mateus Roder\IEEEauthorrefmark{1}, Jurandy Almeida\IEEEauthorrefmark{2}, Gustavo H. de Rosa\IEEEauthorrefmark{1}, Leandro A. Passos\IEEEauthorrefmark{1}, \\Andr\'e L. D. Rossi\IEEEauthorrefmark{1}, Jo\~ao P. Papa\IEEEauthorrefmark{1}}
\IEEEauthorblockA{\IEEEauthorrefmark{1}Department of Computing, S\~ao Paulo State University -- UNESP, Bauru, Brazil \\
\{mateus.roder, gustavo.rosa, leandro.passos, andre.rossi, joao.papa\}@unesp.br}
\IEEEauthorblockA{\IEEEauthorrefmark{2}Instituto de Ci\^encia e Tecnologia, Universidade Federal de S\~ao Paulo -- UNIFESP, S\~ao Jos\'e dos Campos, Brazil \\
jurandy.almeida@unifesp.br}
}


%

\else
  \author{Sibgrapi paper ID: \cmtid \\ }
\fi

\maketitle

\begin{abstract}
In the last decade, exponential data growth supplied machine learning-based algorithms' capacity and enabled their usage in daily-life activities. Additionally, such an improvement is partially explained due to the advent of deep learning techniques, i.e., stacks of simple architectures that end up in more complex models. Although both factors produce outstanding results, they also pose drawbacks regarding the learning process as training complex models over large datasets are expensive and time-consuming. Such a problem is even more evident when dealing with video analysis. Some works have considered transfer learning or domain adaptation, i.e., approaches that map the knowledge from one domain to another, to ease the training burden, yet most of them operate over individual or small blocks of frames. This paper proposes a novel approach to map the knowledge from action recognition to event recognition using an energy-based model, denoted as Spectral Deep Belief Network. Such a model can process all frames simultaneously, carrying spatial and temporal information through the learning process. The experimental results conducted over two public video dataset, the HMDB-51 and the UCF-101, depict the effectiveness of the proposed model and its reduced computational burden when compared to traditional energy-based models, such as Restricted Boltzmann Machines and Deep Belief Networks.
\end{abstract}


\IEEEpeerreviewmaketitle

\section{Introduction}
\label{s.intro}

Machine Learning (ML) techniques emerged in the last decades as revolutionary tools capable of solving or slightly alleviating the burden imposed by repetitive and tedious tasks. Recently, the advent of Deep Learning (DL) algorithms powered up such advances, providing astonishing predictions over complex domains. On the other hand, video-based domain tasks still pose challenging assignments to intelligent algorithms, e.g., recognizing human actions in videos~\cite{YuArxiv:18}.

An action recognition task is characterized by an observation of a complete sequence of movements performed by a human followed by its classification~\cite{bobick2001recognition}. Such a task plays a fundamental role in video surveillance-based security systems, content-based video retrieval, and self-driving cars, among others. Event recognition is a similar task which models the ability to retrieve specific actions from sequences of videos, focusing on learning behaviors related to events of interest, i.e., events which comprise specific activities and objects in a given scene. Thus, there is a slight difference between action and event recognition tasks, i.e., while the former attempts to identify any action performed in the scene, the latter remarks specific movements, such as ``is there anybody drinking in the street?'' or ``does this video contains any unusual behavior?''. Event recognition approaches are commonly employed for monitoring public or private areas in search for anomalous behaviors, such as violent assaults on sports events, abandoned objects on train stations, and route obstruction on industrial environments, among others.

Despite their similarities, both action and event recognition tasks present their particularities, posing distinct challenges while training an intelligent model. An interesting approach used to solve event recognition tasks is importing some correlated knowledge extracted over action recognition-based models, denoted as transfer learning. 

Transfer learning studies the possibilities of transferring knowledge from source domains to different contexts (target domains). To illustrate such an idea, consider an autonomous-driving car trained with road-traffic data from New Zealand. It will not work effectively on Brazilian streets due to different signage and road-traffic rules, a distinct influx of vehicles, and right-lane-based driving, among other issues. However, adapting the knowledge learned in New Zealand to Brazil can reduce computational costs and time needed to train new models~\cite{venkateswara2017deep}. In a nutshell, the transfer learning considers two or more feature space distributions been related or equal, where the task related to each one can be also related or equal, providing useful information for the target task~\cite{sun2015survey}.

Several works addressed the problem of action and event recognition through transfer learning strategies. Farajidavar et al.~\cite{farajidavar2011transductive}, for instance, proposed a transductive transfer learning method for action recognition in tennis games. Further, Shao et al.~\cite{shao2014transfer} published a survey comprising several approaches for these tasks, such as space approximation~\cite{quattoni2008transfer}, Gaussian mixture model~\cite{cao2010cross}, and geometric reasoning~\cite{darrell1996task}. Recently, novel approaches considered DL methods, such as Wang et al.~\cite{wang2018pm}, who introduced Generative Adversarial Networks for action recognition using partial-modalities. Tas et al.~\cite{tas2018cnn} employed a Convolutional Neural Network (CNN) for action recognition and supervised domain adaptation on 3D body skeletons. Furthermore, Gao et al.~\cite{gao2019know} introduced a two-stream graph CNN for zero-shot action recognition, while Liu et al.~\cite{liu2019deep} proposed image-to-video adaptation and fusion networks in the same context.  

Among several DL algorithms, an energy-based model known as Deep Belief Network (DBN)~\cite{Hinton:06} obtained considerable popularity in the last years due to its notable results in a wide variety of applications~\cite{ali2018speaker, RoderICAISC:20}. It is composed of multiple hidden layers, such that each layer is a greedily trained Restricted Boltzmann Machine (RBMs)~\cite{Hinton:02}. Even though some works proposed DBNs for action~\cite{gowda2017human,ali2014learning} and a specific type of event recognition~\cite{zhang2017event}, as far as we know, there is still no work that has successfully addressed the concepts of transfer learning from actions to events with video through DBNs, i.e., to learn useful features from highly structured actions and movements to the generalization of high-level events. Therefore, the main contributions of this work are threefold: (i) to introduce DBNs in the video event classification domain, (ii) to propose two approaches, denoted as Aggregative-DBN and Gradient-DBN, that employ frame fusion and image gradient respectively, and (iii) to support the lack of video-based event recognition in literature.

Additionally, we consider the following hypotheses: (i) DBNs are able to learn useful correlations that map actions to high-level events in video-based domain tasks; (ii) the overall accuracy can be improved using the proposed approaches; and (iii) the overall training time can be reduced with the Aggregative-based approach.

The remainder of this paper is organized as follows. Section~\ref{s.theoretical} introduces the main theoretical concepts used in the manuscript, while Section~\ref{s.proposed} presents the proposed approaches. Further, Sections~\ref{s.methodology} and~\ref{s.experiments} describe the methodology adopted in this work and the experimental results. Finally, Section~\ref{s.conclusion} states conclusions and future work.

\section{Theoretical Background}
\label{s.theoretical}


\subsection{Video-based Domain}
\label{ss.video}

Let $\mathcal{F} = \{F_{1}, F_{2}, \dots, F_n\}$ be a temporal sequence of frames (images) $F_{i}$, which represents a soundless scene and possible movements of its components. Such frames can be classified according to the complexity of their internal representations and the interaction level between their entities. Thus, it is possible to generate four classification-based categories: attributes and movements, low-level events and actions, interaction, and high-level events~\cite{Jiang2013}.

\begin{figure}[!ht]
    \centering
    \includegraphics[scale=0.3]{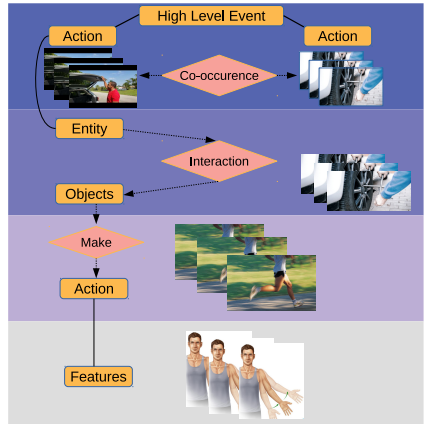}
    \caption{Video-based domain hierarchical complexity.}
    \label{f.hierarchy}
\end{figure}

Figure~\ref{f.hierarchy} illustrates the hierarchical form of the categories discussed hereafter. The movement characterizes the lowest representation level of a frame and is widely employed to recognize human actions, such as body movement~\cite{Liu:11}. On the other hand, low-level events and actions represent a particular chain of movements, usually carried out by an entity, e.g., car or person. Additionally, if such actions are performed or interacted by more than one entity or object, it is possible to categorize them as an interaction~\cite{Jiang2013}. Finally, the highest-level category, denoted as complex events, represents the interaction of entities or sequence of actions in a specific time window in the video. For instance, one can identify a birthday party as an event composed of several actions and entities in a single scene. 

Therefore, the event recognition task attempts to detect the complex events' spatial and temporal locations in a sequence of frames~\cite{Jiang2013}. The literature lacks in standardizing the difference between actions and events, making them interchangeable in most applications~\cite{Jiang2013}. 

\subsection{Applied Transfer Learning}
\label{ss.d_transfer}

Transfer learning has recently received attention due to the advent of the ImageNet\footnote{https://image-net.org/} dataset and the increased processing power of GPUs. Such a task consists in transferring the knowledge from the source domain to the target domain, providing information that can be useful on the target domain, mainly when data are insufficient or the computational resources are scarce. In this way, it is possible to train deep neural networks in large-scale image/video datasets and use them to fine-tune more specific tasks~\cite{tan2018survey}.

Given the previous concepts, it is possible to elucidate the mathematical formulation regarding the problem addressed here. Let $\Gamma$ be a high-level event recognition task, as well as let $\mathcal{D}_{S}$ and $\mathcal{D}_{T}$ be the source (action domain) and the target space domains (event domain), respectively. Additionally, the source domain is composed of the subspaces $\mathcal{A} \in \mathbb{R}^{d_a}$, $\mathcal{M} \in \mathbb{R}^{d_m}$, and $\mathcal{I} \in \mathbb{R}^{d_i}$, where $\{\mathcal{A}, \mathcal{M}, \mathcal{I}\}\subset \mathcal{D}_{S}$, while the target domain is composed of $\mathcal{E} \in \mathbb{R}^{d_e}$, where $\{\mathcal{E}\} \subset \mathcal{D}_{T}$. The subspace $\mathcal{A}$ stands for the $d_a$-dimensional base actions, $\mathcal{M}$ stands for the $d_m$-dimensional movements, $\mathcal{I}$ represents the interactions between $d_i$-dimensional entities, and $\mathcal{E}$ stands for the $d_e$-dimensional high-level events. 

From the transfer learning theory~\cite{weiss2016survey}, a specific case is when the data from the source domain and target domain keep their probabilities, letting $\mathcal{D}_{S}$ to be identical to $\mathcal{D}_{T}$, differing only on the target task, which is the problem addressed in this paper. Finally, it is possible to formulate the proposed approach for the given task using Equation~\ref{e.transfer}, as follows:

\begin{equation}
\label{e.transfer}
\Gamma = \{y_{\mathcal{D}_{T}}, f(\mathcal{D}_{S}) \},
\end{equation}
where $y_{\mathcal{D}_{T}}$ stands for the target domain labels and $f(\mathcal{D}_{S})$ for the function that learns features from the source domain. Such learned features are useful for the target domain and its respective event classification task, i.e., a neural network that learns from $\mathcal{D}_{S}$ and is fine-tuned in $\mathcal{D}_{T}$ with labels $y_{\mathcal{D}_{T}}$. Here, $\mathcal{D}_{T}$ has the same probability distribution that $\mathcal{D}_{S}$, as aforementioned.

\subsection{Restricted Boltzmann Machines}
\label{ss.rbm}

Restricted Boltzmann Machines (RBMs)~\cite{Smolensky:86,Hinton:02} are described as a bipartite graph composed of two layers of neurons, i.e., a visible layer $\textbf{v}\in\{0,1\}^m$, which is responsible for the input data, and a hidden layer $\textbf{h}\in\{0,1\}^n$, whose units map the data representation into a latent space. This interaction is modeled by a weight matrix, $\textbf{W}\in\Re^{m\times n}$, which connects each visible unit $v_i$ to all hidden units $h_j$, and vice-versa, denoted by the arc $w_{ij}$. 

The RBM learning procedure is performed by the minimization of an energy function concerning some intrinsic variables, described as follows:

\begin{equation}
\label{e.energy_bbrbm}
	E(\textbf{v},\textbf{h})=-\sum_{i=1}^mb_iv_i-\sum_{j=1}^nc_jh_j-\sum_{i=1}^m\sum_{j=1}^nv_ih_jw_{ij},
\end{equation}
where $\textbf{b}\in\Re^m$ and $\textbf{c}\in\Re^n$ stand for the bias vector considering the visible and hidden layers, respectively.

Computing the joint probability of the system poses an intractable task due to the increasing number of possible states. However, since the model is represented as a bipartite graph, one can compute both the visible and hidden units' activation in a mutually independent fashion, performed as follows:

\begin{equation}
\label{e.probv}
	P(v_i=1|\textbf{h})=\phi\left(\sum_{j=1}^nw_{ij}h_j+b_i\right),
\end{equation}

and

\begin{equation}
\label{e.probh}
	P(h_j=1|\textbf{v})=\phi\left(\sum_{i=1}^mw_{ij}v_i+c_j\right).
\end{equation}
Note that $\phi(\cdot)$ stands for the logistic-sigmoid function. Finally, we can solve the equations above by iteratively sampling over a Markov Chain, using the well-known Contrastive Divergence (CD) algorithm~\cite{Hinton:02}.

The learning process in RBMs consists of an optimization problem whose goal is to minimize the energy function given in Equation~\ref{e.energy_bbrbm}. In other words, such a process ends up maximizing the marginal probability distribution of the visible units, defined as follows:

\begin{equation}
\label{e.proba}
P(\textbf{v})=\frac{\displaystyle\sum_{\textbf{h}}e^{-E(\textbf{v},\textbf{h})}}{\displaystyle\sum_{\textbf{v},\textbf{h}}e^{-E(\textbf{v},\textbf{h})}},
\end{equation}
which is commonly handled in its natural logarithm version, i.e., more precisely, we aim at maximizing the negative logarithm of the likelihood function (Negative Log-Likelihood - NLL). Moreover, regarding the visible units, such a procedure can be easily extended to the continuous domain, which is useful to model any type of input. The changes occur on the energy function, as follows:

\begin{equation}
\label{e.energy_gbrbm}
E(\textbf{v},\textbf{h})= \sum_{i=1}^m \dfrac{(v_i-b_i)^2}{2\sigma^{2}_{i}} - \sum_{j=1}^nc_jh_j - \sum_{i=1}^m\sum_{j=1}^n\dfrac{v_i}{\sigma_i}h_jw_{ij}.
\end{equation}

Considering the derivatives, it is straightforward to show that the visible prior becomes:

\begin{equation}
\label{e.gaussv}
P(v_i=1|\textbf{h}) \sim\mathcal{N} \left(\sum_{j=1}^nw_{ij}h_j+b_i, \sigma^{2}_{i}\right),
\end{equation}
and, when a Gaussian input with zero mean and one unit standard deviation is employed, such a prior becomes simple and easy to sample, since the whole procedure is still the same.

\subsection{Deep Belief Networks}
\label{ss.dbn}

Deep Belief Networks (DBNs)~\cite{Hinton:06} are generative graphical models composed of a stack of RBMs, thus providing multiple layers of latent variables, such that the hidden layer of the bottommost RBM is employed to feed the subsequent input units successively until reaching the topmost layer. DBNs are trained greedily, meaning that an RBM at a specific layer does not consider others during its learning procedure. Thus, a DBN is composed of $L$ layers, where $\textbf{W}^l$ is the weight matrix of an RBM at layer $l$. Additionally, we can observe that the hidden units at layer $l$ become the input units of layer $l+1$. 

The aforementioned procedure stands for the generative pre-training. Afterward, it is possible to attach fully-connected (FC) layers with softmax outputs at the topmost hidden layer for a discriminative fine-tuning, which can be used for classification tasks.

\section{Proposed Approaches}
\label{s.proposed}

This work proposes to employ DBNs as non-linear functions $f(\mathcal{D}_{S})$ to learn from the source domain, $\mathcal{D}_{S}$, the information that can be used to map the target domain, $\mathcal{D}_{T}$, i.e., to extract information from videos and use them to classify high-level events. Additionally, two approaches are shown to the respective task.

\subsection{Aggregative Deep Belief Networks}
\label{ss.sdn}

The first alternative architecture is denoted as Aggregative (A- prefix in models), which modifies the DBNs' first layer. Such a variation is designed considering two main concepts: (i) to capture general spatio-temporal information without additional techniques, such as optical flow algorithms, and (ii) reduce the overall computational burden.

\begin{figure}[!ht]
    \centering
    \includegraphics[scale=.4]{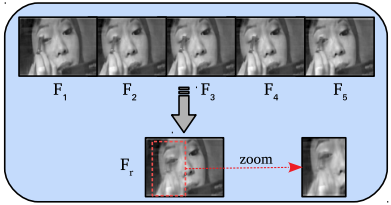}
    \caption{Frames aggregation for the Aggregative-based approach.}
    \label{f.frames}
\end{figure}

The proposed A-DBN processes all frames simultaneously instead of processing one frame at a time, enabling a complete parameter update at each iteration. In other words, A-DBN aggregates all frames $\{F_{1}, F_{2}, \dots, F_n\}$ into a single frame denoted as $F_r$, which represents their summation. This procedure carries spatial information (contour and edges) along their temporal trajectories, highlighted as ``spectrum'' in the resulting frame. Figure~\ref{f.frames} depicts the $F_r$ aggregation process, as well as its highlighted  region. Afterward, A-DBN first layer infers the posterior distribution given $F_r$, as follows:

\begin{equation}
\label{e.sdbn_dist}
P(h_j=1|\bm{F_{r}})=\phi\left(\sum_{i=1}^mw_{ij}F_{ri}+c_j\right).
\end{equation}

\subsection{Gradient Deep Belief Networks}
\label{ss.gdbn}

The second alternative architecture denoted as Gradient (G- prefix in models) also modifies the DBNs first layer. Such a variation is designed considering one main concept, i.e., to capture general motion information between two frames.

\begin{figure}[!ht]
    \centering
    \includegraphics[scale=.4]{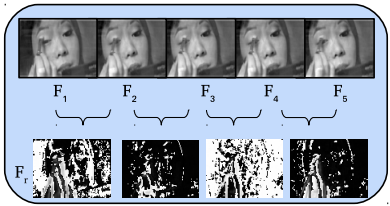}
    \caption{Frames generation for the Gradient-based approach.}
    \label{f.frames2}
\end{figure}

The proposed G-DBN processes two consecutive frames instead of processing one frame at a time. In other words, G-DBN defines a resultant frame, $F_r$, as the direct subtraction as follows: given $F_{1}$ and $F_{2}$, $F_r$ stands for $F_{2} - F_{1}$. This procedure carries motion cues from the spatial domain along their trajectories. Figure~\ref{f.frames2} depicts the $F_r$  generation process, as well as its highlighted region. Afterward, G-DBN first layer infers the posterior distribution given $F_r$, as follows:

\begin{equation}
\label{e.sdbn_dist2}
P(h_j=1|\bm{F_{r}})=\phi\left(\sum_{i=1}^mw_{ij}F_{ri}+c_j\right).
\end{equation}

\section{Experiments}
\label{s.methodology}

In this section, we describe the dataset and the experimental setup employed to apply and compare DBNs with A-DBN, proposed in this paper, for the task of event recognition from complex actions domain.

\subsection{Dataset}
\label{ss.dataset}

We opted to use two well-known datasets, UCF-101~\cite{soomro2012ucf101} and HMDB-51~\cite{Kuehne2011HMDB}, as they represent a big challenge and are well-established video action recognition datasets. Both datasets comprises a significant amount of data from real-world action videos collected from YouTube and classified among $101$ and $51$ distinct classes, respectively. Moreover, such a diversity becomes more expressive due to the substantial variations in camera motion, object appearance and pose, scaling, viewpoint, cluttered background, and illumination conditions.

\begin{figure}[!ht]
    \centering
    \includegraphics[scale=0.38]{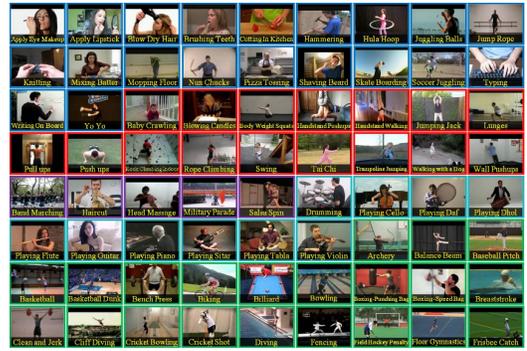}
    \caption{Random clips from UCF-101~\cite{soomro2012ucf101}.}
    \label{f.ucf}
\end{figure}

The $13,320$ videos from UCF-101 are grouped in 5 macro-categories, which are easily interpreted as high-level events. Also, this mapping expects inter-class videos to share standard and essential features, which helps in recognizing common actions and interactions. Following the authors' guideline, the high-level events used: (0) sports practice; (1) musical practice with an instrument; (2) human-object interaction; (3) human body-motion; and (4) people interacting. Random clips depicting such classes are presented in Figure~\ref{f.ucf}, each class is represented by the color of the border: green for $0$, light-blue for $1$, blue for $2$, red for $3$, and purple for $4$.

\begin{figure}[!ht]
    \centering
    \includegraphics[scale=0.38]{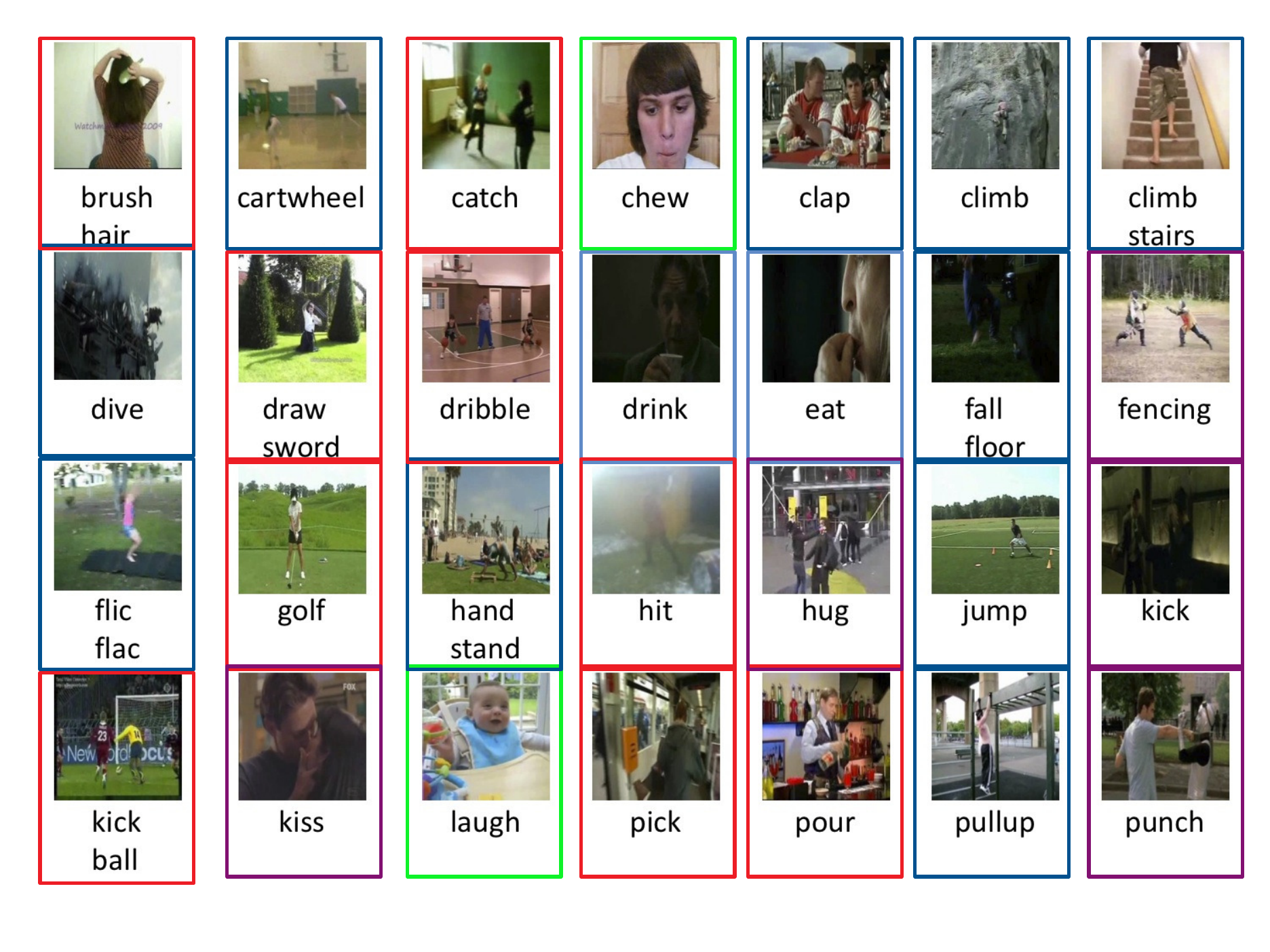}
    \caption{Random clips from HMDB-51~\cite{Kuehne2011HMDB}.}    
    \label{f.hmdb}
\end{figure}

Similarly, the $6,766$ videos from HMDB-51 are grouped in 5 macro-categories, easily interpreted as high-level events. Also, following the authors' guideline, the high-level events are (0) human facial expression; (1) manipulation of objects in the face region; (2) body movement; (3) interaction between people and object(s); and (4) person interacting with each other, where numbers in parentheses represent classes. Additionally, frames of clips from the dataset are shown in Figure~\ref{f.hmdb}, where the color of the border represents the event class: green for $0$, light blue for $1$, blue for $2$, red for $3$, and purple for $4$.

Both datasets provide three partitions, each with separate data for training and testing, where the first partition was used in this study, as it seems to have the most difficult test samples to evaluate as cluttered background or fewer interactions and actions. The process of splitting/acquiring the frames is performed in a similar way to the work of Ng et al.~\cite{yue2015beyond}, using $6$ frames per video clip uniformly distributed over time. In their work, Ng et al. showed that $6$ frames per video are enough to ensure a good performance, achieving the same results as $20$ frames, for example, in addition to imprinting a lower computational load on the action classification task.

Regarding the pre-processing step, two transformations were employed before the image conversion to grayscale. The first transformation concerns cutting operations, removing black regions that do not carry information, and resizing from the original size ($240 \times 320$) to $72 \times 96$, to facilitate the processing of energy-based models. The second transform stands for feature-normalization using a Gaussian distribution with zero mean and unit variance.

\subsection{Experimental Setup}
\label{ss.experimental}

Regarding the experimental setup hardware, we employed an Intel 2x Xeon(R) E5-2620 @ 2.20GHz (40 cores), a GTX 1080 Ti, and 128 GB of RAM. For the unsupervised pre-training process, we opted to use mini-batches of $128$ samples and $3$ epochs per-layer. Finally, Table~\ref{t.hyper} describes the employed architectures and hyper-parameters.

\begin{table*}[!ht]
    \renewcommand{\arraystretch}{1.1}
    \centering
    \caption{Configuration of the models used in this work.}
    \label{t.hyper}
    \begin{tabular}{lclll}
        \toprule
        \textbf{Model} & \textbf{Layers} & \textbf{Hidden Neurons} & \textbf{Momentum} & \textbf{Learning Rate}
        \\ \midrule
        RBM & $ 1 $&$ [2,000] $ & $ [0.5] $ & $ [1\cdot10^{-3}] $ \\ \midrule
        A-RBM & $ 1 $&$ [2,000] $ & $ [0.5] $ & $ [1\cdot10^{-3}] $ \\ \midrule
        G-RBM & $ 1 $&$ [2,000] $ & $ [0.5] $ & $ [1\cdot10^{-3}] $ \\ \midrule
        $\text{DBN}_{\alpha}$ & $ 2 $&$ [2,000-2,000]$ & $ [0.5; 0.5] $ & $ [1\cdot10^{-3}; 5\cdot10^{-4}] $ \\ \midrule
        $\text{DBN}_{\beta}$  & $ 3 $&$ [2,000-2,000-2,000]$ & $ [0.5; 0.5; 0.5] $ & $ [1\cdot10^{-3}; 5\cdot10^{-4}; 5\cdot10^{-4}] $ \\ \midrule
        $\text{DBN}_{\iota}$  & $ 2 $&$ [4,000-4,000]$ & $ [0.5; 0.5] $ & $ [5\cdot10^{-4}; 5\cdot10^{-4}] $ \\ \midrule
        $\text{DBN}_{\zeta}$  & $ 3 $&$ [4,000-4,000-4,000]$ & $ [0.5; 0.5; 0.5] $ & $ [5\cdot10^{-4}; 5\cdot10^{-4}; 5\cdot10^{-4}] $ \\ \midrule
        $\text{A-DBN}_{\alpha}$  & $ 2 $&$ [2,000-2,000] $ & $ [0.5; 0.5] $ & $ [1\cdot10^{-3}; 5\cdot10^{-4}] $ \\ \midrule
        $\text{A-DBN}_{\beta}$ & $ 3 $&$ [2,000-2,000-2,000] $ & $ [0.5; 0.5; 0.5] $ & $ [1\cdot10^{-3}; 5\cdot10^{-4}; 5\cdot10^{-4}] $ \\ \midrule
        $\text{A-DBN}_{\iota}$ & $ 2 $&$ [4,000-4,000] $ & $ [0.5; 0.5] $ & $ [5\cdot10^{-4}; 5\cdot10^{-4}] $ \\ \midrule
        $\text{A-DBN}_{\zeta}$ & $ 3 $&$ [4,000-4,000-4,000] $ & $ [0.5; 0.5; 0.5] $ & $ [5\cdot10^{-4}; 5\cdot10^{-4}; 5\cdot10^{-4}] $ \\ \midrule

		$\text{G-DBN}_{\alpha}$  & $ 2 $&$ [2,000-2,000] $ & $ [0.5; 0.5] $ & $ [1\cdot10^{-3}; 5\cdot10^{-4}] $ \\ \midrule
        $\text{G-DBN}_{\beta}$ & $ 3 $&$ [2,000-2,000-2,000] $ & $ [0.5; 0.5; 0.5] $ & $ [1\cdot10^{-3}; 5\cdot10^{-4}; 5\cdot10^{-4}] $ \\ \midrule
        $\text{G-DBN}_{\iota}$ & $ 2 $&$ [4,000-4,000] $ & $ [0.5; 0.5] $ & $ [5\cdot10^{-4}; 5\cdot10^{-4}] $ \\ \midrule
        $\text{G-DBN}_{\zeta}$ & $ 3 $&$ [4,000-4,000-4,000] $ & $ [0.5; 0.5; 0.5] $ & $ [5\cdot10^{-4}; 5\cdot10^{-4}; 5\cdot10^{-4}] $ \\ \bottomrule        
    \end{tabular}
\end{table*}

Each model is connected to two additional fully-connected layers that are fine-tuned using the well-known Adam optimizer~\cite{Kingma:14} with the learning rate equals to $10^{-3}$ and the same number of epochs and mini-batch size. Such FC layers have two configurations since they depend on the number of hidden neurons from the last RBM/DBN layer, i.e., $2,000-1,000-5$ and $4,000-2,000-5$. It is important to highlight that the RBMs were employed for $2,000$ hidden neurons only as this work primarily focuses on using the hierarchical information learned by DBNs.

Furthermore, we opted to use the same approach employed by transfer learning, i.e., to freeze the connections from the first hidden layer and make a gentle adjustment of subsequent hidden layers (learning rate equals to $10^{-6}$). The cross-entropy loss was used when adjusting model weights during the fine-tuning process, while the final measure was the accuracy on the testing set. Finally, to mitigate any stochastic nature, each model was fully trained and fine-tuned for $6$ repetitions.

\section{Experimental Results}
\label{s.experiments}

This section presents the experimental results concerning the DBNs and the proposed approaches, i.e., the A-DBN and the G-DBN, applied to the task of event recognition. All models follow the work of Ng et al.~\cite{yue2015beyond}, using $6$ frames per video clip uniformly distributed over time.

\subsection{Model Evaluation for Event Recognition}
\label{ss.classification}

Regarding the main task, i.e., learning from actions to classify high-level events, Tables~\ref{t.accuracy_ucf} and \ref{t.accuracy_hmdb} show the predictive performance for all models and architectures. The highlighted result stands for the best mean accuracy. Besides, it also presents the average running time for each model, averaged over the six repetitions to analyze the models' computation impact and efficiency. 

\begin{table}[!ht]
    \renewcommand{\arraystretch}{1.25}
    \centering
    \caption{Mean accuracies (\%) and running times (minutes) over the UCF-101 test set (fold 1).}
    \label{t.accuracy_ucf}
    \begin{tabular}{lcc}
    	\toprule
    	\textbf{Architecture} & \textbf{Accuracy} & \textbf{Time}
    	\\ \midrule
    	RBM & $38.71 \pm 1.04$ & $315.00 \pm 5.00$ \\ \midrule
    	A-RBM & $42.48 \pm 0.94$ & $270.00 \pm 5.00$ \\ \midrule    	
    	G-RBM & $44.04 \pm 1.82$ & $314.00 \pm 5.00$ \\ \midrule
    	
    	$\text{DBN}_{\alpha}$ & $37.72 \pm 4.67$ & $765.00 \pm 5.00$ \\ \midrule
    	$\text{A-DBN}_{\alpha}$ & $44.66 \pm 1.28$ & $540.00 \pm 5.00$ \\ \midrule
    	$\text{G-DBN}_{\alpha}$ & $40.16 \pm 11.63$ & $764.00 \pm 5.00$ \\ \midrule
    	
    	$\text{DBN}_{\beta}$ & $40.55 \pm 3.54$ & $1,215.00 \pm 5.00$ \\ \midrule
    	$\text{A-DBN}_{\beta}$ & $44.80 \pm 2.02$ & $810.00 \pm 5.00$ \\ \midrule
    	$\text{G-DBN}_{\beta}$ & $44.84 \pm 0.08$ & $1,211.00 \pm 5.00$ \\ \midrule    	
    	
    	$\text{DBN}_{\iota}$ & $41.92 \pm 2.65$ & $775.00 \pm 6.00$ \\ \midrule
    	$\text{A-DBN}_{\iota}$ & $\bm{45.01 \pm 1.39}$ & $550.00 \pm 6.00$ \\ \midrule
    	$\text{G-DBN}_{\iota}$ & $44.84 \pm 0.04$ & $773.00 \pm 6.00$ \\ \midrule    	    	
    	
    	$\text{DBN}_{\zeta}$ & $42.33 \pm 4.51$ & $1,225.00 \pm 6.00$ \\ \midrule
    	$\text{A-DBN}_{\zeta}$ & $44.87 \pm 2.81$ & $820.00 \pm 6.00$ \\ \midrule
    	$\text{G-DBN}_{\zeta}$ & $44.86 \pm 0.14$ & $1,220.00 \pm 6.00$ \\ \bottomrule
    	
    \end{tabular}
\end{table}

From Table~\ref{t.accuracy_ucf}, we can notice prominent results, mainly for the proposed Aggregative version of DBNs. Starting from the base model, A-RBM achieved a better accuracy than RBM, i.e., $42.48\%$ against $38.71\%$, representing a meaningful difference ($3.77$ points of mean percentual accuracy), while the G-RBM model achieved $44.04\%$, outperforming also the Aggregative approach. In addition, A-RBM has a significantly lower computational burden, approximately $14\%$ less running time. To clarify, hereafter, the percentual difference between models' accuracy stands for the absolute mean value of the proposed model (A- or G-) subtracted to its standard version, and for the running time, it is the mean proposed approach time divided by its standard model.

Regarding the second architecture, i.e., the $\alpha$ models, the Aggregative version overcomes the standard DBN and G-DBN in mean accuracy by almost $7\%$ and $4.5\%$, respectively, representing a meaningful improvement. However, the standard model and the G-DBN do not overpass its parameterless version (RBM and G-RBM) in mean accuracy. Moreover, the running time for A-DBN$_{\alpha}$ was $30\%$ smaller than DBN$_{\alpha}$, impacting positively to a lighter training burden.

Concerning the $\beta$ models, a similar behavior was observed, i.e., the Aggregative version overpassed the standard DBN in approximately $4\%$ in mean accuracy, and almost $33\%$ less running time. Also, the G-DBN model achieved the best mean accuracy, $44.84\%$, with a small standard deviation. However, such results show that adding more hidden layers does not lead to an impressive performance improvement for A-DBN since its mean accuracy was close to the A-DBN$_\alpha$, while for the G-DBN$_\beta$ the performance was increased. Moreover, even with the previous observation, the A-DBN$_\beta$ still gives a better running time than its baselines.

Regarding the fourth architecture ($\iota$ models), the same behavior was observed, highlighting that A-DBN$_{\iota}$ achieved a remarkable mean accuracy of $45.01\%$, the highest average value over the baselines. Moreover, the running time for A-DBN$_{\iota}$ was $30\%$ better than its standard version. Here, the Aggregative models showed that more hidden units might benefit the overall performance for the event classification task with transfer learning.

Finally, the $\zeta$ models showed almost the same results as the $\iota$ models, mainly for the proposed approaches, A-DBN and G-DBN, which achieved a mean accuracy of $44.87\%$ and $44.86\%$, respectively. Such results indicate that more hidden neurons improve the models' performance. However, the larger versions may demand more epochs of pre-training and/or data. Also, the running time of A-DBN overpasses its standard and the Gradient version by approximately $33\%$.

Nonetheless, it is essential to notice that A-DBN models have no difficulty in overpassing the A-RBM mean accuracy; however, the G-DBN$_\alpha$ do not overpass its simpler version on mean accuracy due to one specific run that pushed down the performance (note the standard deviation). Overall, the performance improvement observed can be directly linked to the higher abstraction achieved by hidden layers. Such approaches can improve the DBN lower bound and provide a further improvement in discriminative fine-tuning. Besides, they were also pre-trained with a relatively small number of epochs, which can induce a less efficient overall lower bound optimization. However, the results showed that more hidden units in hidden layers improve the mean accuracy rate, as the A-DBN$_{\iota}$ and A-DBN$_{\zeta}$ models have shown.

\begin{table}[!ht]
    \renewcommand{\arraystretch}{1.25}
    \centering
    \caption{Mean accuracies (\%) and running times (minutes) over the HMDB-51 test set (fold 1).}
    \label{t.accuracy_hmdb}
    \begin{tabular}{lcc}
    	\toprule
    	\textbf{Architecture} & \textbf{Accuracy} & \textbf{Time}
    	\\ \midrule
    	RBM & $34.60 \pm 3.90$ & $45.00 \pm 5.00$ \\ \midrule
    	A-RBM & $35.19 \pm 4.24$ & $30.00 \pm 5.00$ \\ \midrule
    	G-RBM & $38.05 \pm 0.36$ & $44.00 \pm 5.00$ \\ \midrule
    	    	
    	$\text{DBN}_{\alpha}$ & $34.49 \pm 3.98$ & $144.00 \pm 5.00$ \\ \midrule
    	$\text{A-DBN}_{\alpha}$ & $37.68 \pm 4.19$ & $132.00 \pm 5.00$ \\ \midrule
    	$\text{G-DBN}_{\alpha}$ & $38.40 \pm 0.01$ & $143.00 \pm 5.00$ \\ \midrule
    	    	
    	$\text{DBN}_{\beta}$ & $33.83 \pm 4.06$ & $216.00 \pm 5.00$ \\ \midrule
    	$\text{A-DBN}_{\beta}$ & $38.70 \pm 4.33$ & $195.00 \pm 5.00$ \\ \midrule
    	$\text{G-DBN}_{\beta}$ & $38.40 \pm 0.01$ & $214.00 \pm 5.00$ \\ \midrule
    	
    	$\text{DBN}_{\iota}$ & $34.41 \pm 4.03$ & $150.00 \pm 6.00$ \\ \midrule
    	$\text{A-DBN}_{\iota}$ & $38.23	\pm 4.25$ & $138.00 \pm 6.00$ \\ \midrule
    	$\text{G-DBN}_{\iota}$ & $38.40 \pm 0.01$ & $148.00 \pm 6.00$ \\ \midrule
    	    	
    	$\text{DBN}_{\zeta}$ & $34.53 \pm 4.04$ & $225.00 \pm 6.00$ \\ \midrule
    	$\text{A-DBN}_{\zeta}$ & $\bm{38.86 \pm 4.26}$ & $207.00 \pm 6.00$ \\ \midrule
	   	$\text{G-DBN}_{\zeta}$ & $37.41 \pm 2.43$ & $223.00 \pm 6.00$ \\ \bottomrule
    \end{tabular}
\end{table}

From Table~\ref{t.accuracy_hmdb}, one can notice interesting results, mainly for the proposed approaches. The base model, A-RBM achieved a better accuracy rate than RBM, i.e., $35.19\%$ against $34.60\%$, representing a meaningful difference ($0.60$ points of mean percentual accuracy), while the G-RBM model achieved $38.05\%$, beating also the Aggregative approach. Here, it is interesting to note that the time difference between models was not so impressive, explained by the low data volume.

Regarding the second architecture, i.e., the $\alpha$ models, the Aggregative version overcomes the standard DBN in mean accuracy by almost $3\%$, while the G-DBN overpassed its standard version in approximately $4\%$, respectively, representing a meaningful improvement. Moreover, it is important to notice that the running time for A-DBN$_{\alpha}$ was $8\%$ smaller than DBN$_{\alpha}$, impacting positively to a lighter training burden.

Concerning the $\beta$ models, was observed a similar behavior, i.e., the Aggregative version overpassed the standard DBN in approximately $5\%$ in mean accuracy, and almost $10\%$ less running time. Also, the G-DBN model overpassed its standard version in approximately $5\%$ of mean accuracy, with a lower standard deviation. However, such results show that adding more hidden layers does not cause an impressive performance improvement for G-DBN since its mean accuracy was close to the G-DBN$_\alpha$, while for the A-DBN$_\beta$ the performance was increased. Moreover, even with the previous observation, the A-DBN$_\beta$ still gives a better running time than its baselines.

Regarding the $\iota$ models, the baseline model slightly increased its performance, however, the G-DBN achieved the highest mean accuracy, $38.40\%$, overpassing the DBN and A-DBN models. The Aggregative version was also better than DBN, with $38.23\%$, and $8\%$ less running time than DBN. However, such results show that adding more hidden layers does not cause an impressive performance improvement for the proposed approach, keeping the hidden neurons in $2,000$ units, since the mean accuracies were close to the A-DBN$_\beta$ and G-DBN$_\beta$. 

Finally, the $\zeta$ models showed an interesting performance improvement regarding the Aggregative approach, which achieved a remarkable mean accuracy of $38.86\%$. On the other hand, G-DBN suffers from a performance decreasing, achieving $37.41\%$ of mean accuracy. This fact pointed out that more hidden neurons improve the Aggregative's performance while keeping the lowest mean time for training ($207$ minutes).

In general, one can observe two interesting behaviors for the proposed approaches, the Aggregative-based models were able to improve the models' performance, resulted from the total frames aggregation that carries general motion information. On the other hand, the Gradient-based models did not improve their performance as the Aggregative-based, explained by the fact that two-by-two frames on the employed datasets may not carry as much information like the overall sum of the six frames.

\section{Conclusions and Future Works}
\label{s.conclusion}

In this paper, we addressed a transfer learning approach on two well-known video datasets, learning from the action- to the event-based domain, through energy-based models, such as RBMs and DBNs. Furthermore, we proposed Aggregative-based and Gradient-based approaches that modify the processing of the frames, simplifying the model's complexity and giving robustness, saving processing time, and improving the model's generalization. Experimental results show that the proposed approach can reduce the computational time by as much as $33\%$ regarding the A-models.

The results were promising since most A-DBN architectures achieved a feasible mean accuracy rate, highlighting the models A-DBN$_{\iota}$ and A-DBN$_{\zeta}$. Also, the Aggregative models showed a meaningful reduction in running time during the unsupervised pre-training phase. The Gradient models showed a stable behavior, varying almost nothing regarding the different architectures. Therefore, these experimental results support our hypothesis that it is possible to transfer the knowledge from actions to events with the employed energy-based models, without complex inputs such as optical flow or convolutions. Finally, one can highlight that by increasing the number of hidden neurons the overall performance improved, pointing out the models' opportunity to extract more information.

Regarding future works, we plan to investigate the effect of combining convolution operators in energy-based models, such as the Convolutional Restricted Boltzmann Machine (CRBM). Additionally, we aim at employing more complex models, such as Deep Boltzmann Machines (DBMs), trained with a more expressive number of epochs. Finally, a future step is also to analyze how the proposed approach can be applied to data augmentation tasks in the context of event classification in videos.


\section*{Acknowledgments}

This research was supported by S\~{a}o Paulo Research Foundation - FAPESP (grants~\#2013/07375-0, \#2014/12236-1, \#2019/07665-4, \#2019/07825-1 and \#2019/02205-5), FAPESP-Microsoft Research Virtual Institute (grant~\#2017/25908-6), and Brazilian National Council for Scientific and Technological Development - CNPq (grants~\#314868/2020-8, \#307066/2017-7 and \#427968/2018-6).



\bibliographystyle{IEEEtran}
\bibliography{references}
%
%


\end{document}